\documentclass[10pt,twocolumn,letterpaper]{article}

\usepackage{iccv}
\usepackage{times}
\usepackage{epsfig}
\usepackage{graphicx}
\usepackage{amsmath}
\usepackage{amssymb}

\usepackage{booktabs}
\usepackage{multirow}
\usepackage{pifont}
\usepackage{ulem}
\usepackage{bbm}
\usepackage{diagbox}

\usepackage[breaklinks=true,bookmarks=false]{hyperref}

\iccvfinalcopy 

\def\iccvPaperID{3660} 

\ificcvfinal\fi

\begin{document}

\title{Neural Image Re-Exposure}

\author{Xinyu Zhang, Hefei Huang, Xu Jia, Dong Wang, Huchuan Lu\\
Dalian University of Technology\\
{\tt\small \{zhangxy71102, huanghf\}@mail.dlut.edu.cn,  \{xjia, wdice, lhchuan\}@dlut.edu.cn}
}

\maketitle
\ificcvfinal\thispagestyle{empty}\fi

\begin{abstract}
The shutter strategy applied to the photo shooting process has significant influence on the quality of captured photograph. Improper shutter may lead to a blurry image, video discontinuity or rolling shutter artifact. Existing works try to provide an independent solution for each issue.
In this work we aim at re-exposing the captured photo in the post-processing, providing a more flexible way to address those issues within a unified framework. 
Specifically, we propose a neural network based image re-exposure framework. 
It consists of an encoder for visual latent space construction, a re-exposure module for aggregating information to neural film with a desired shutter strategy, and a decoder for `developing' neural film to a desired image.
To compensate for information confusion and missing with frames, event stream, which could capture almost continuous brightness change, is leveraged in computing visual latent content.
Both self-attention layers and cross-attention layers are employed in the re-exposure module to promote interaction between neural film and visual latent content and information aggregation to neural film.
The proposed unified image re-exposure framework is evaluated on several shutter-related image recovery tasks and performs favorably against independent state-of-the-art methods. The code will be made publicly available at \url{https://github.com/zhangxydlut/Neural-Image-Re-Exposure}

\end{abstract}

\section{Introduction}

\begin{figure}  
	\centering
	\includegraphics[width=1.05\columnwidth]{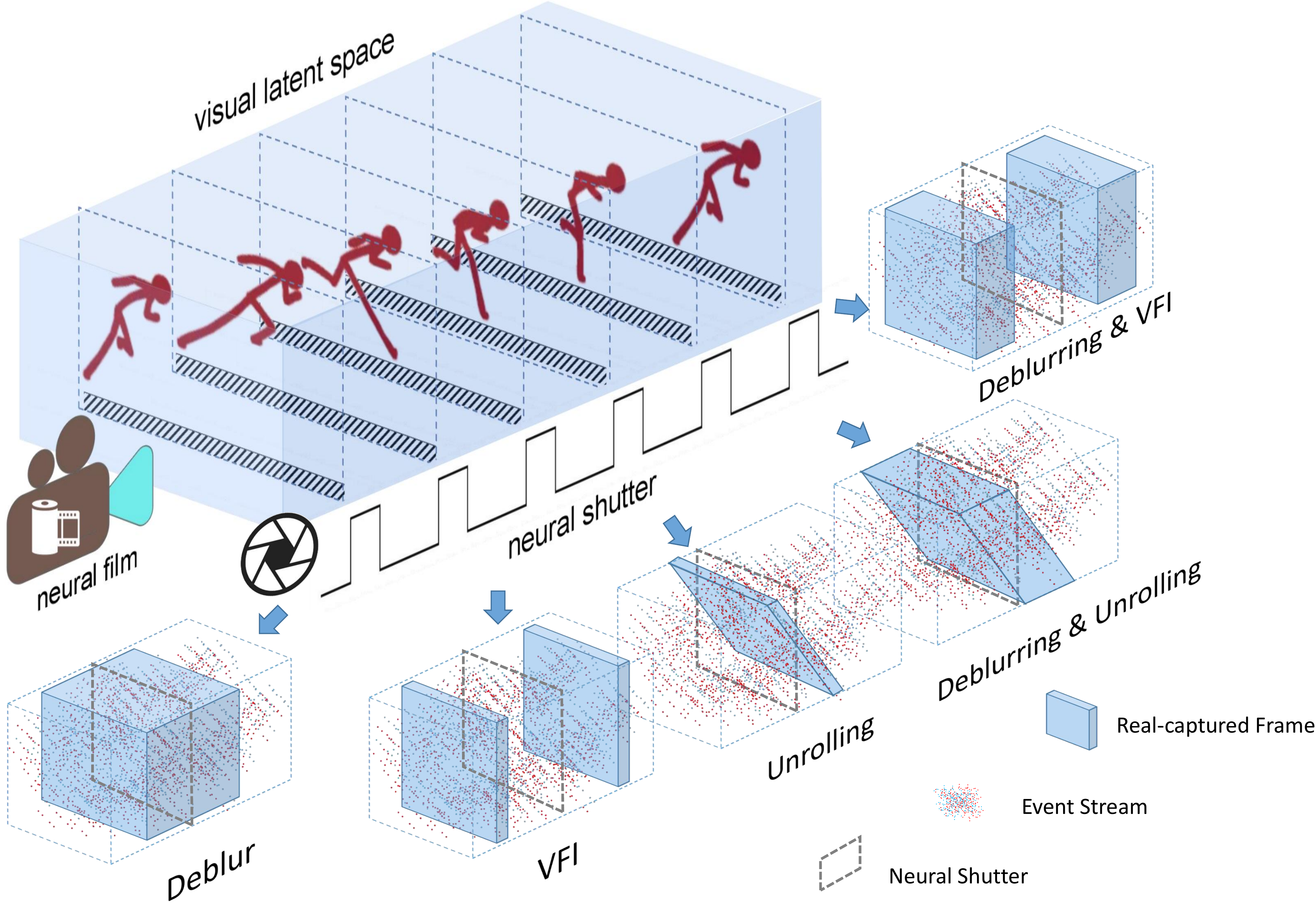}
	\caption{
        Illustration of the proposed framework that could re-expose a captured photo by adjusting the shutter strategy in the post-processing.
	}
	\label{fig:teasing_figure}
\end{figure}

\begin{figure*}[h!]
	\centering
	\includegraphics[width=1.0\textwidth]{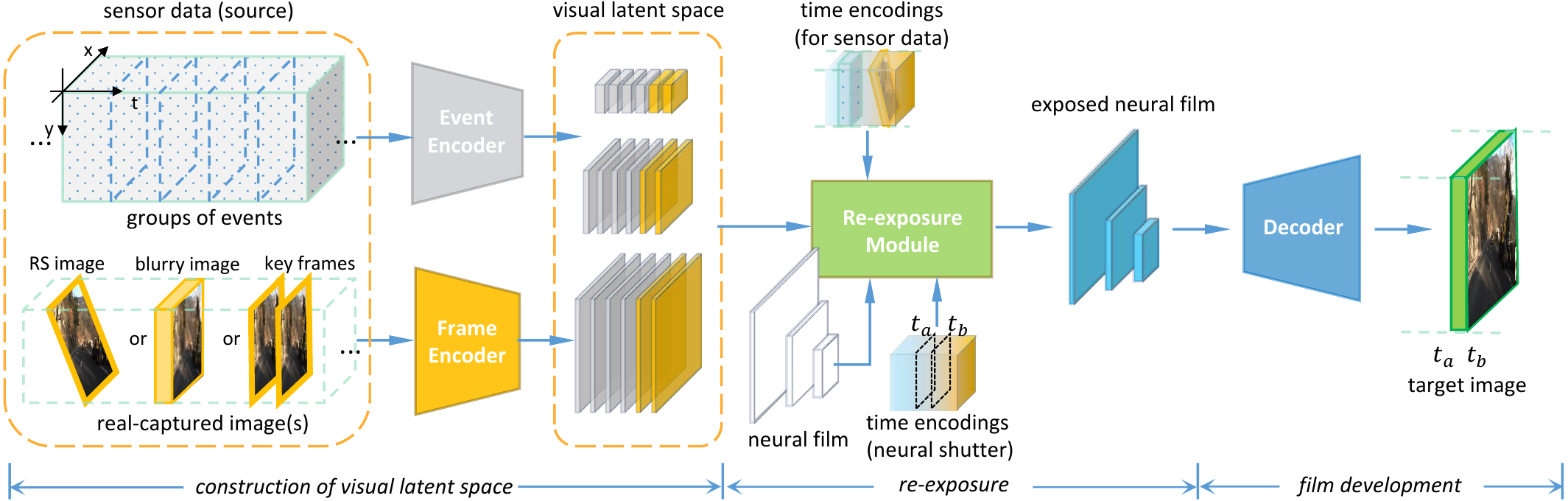}
	\caption{Overall pipeline. 
 By specifying a desired neural shutter,   
 a desired re-exposed image could be produced.
	}
	\label{fig:structure_overview}
	\vspace{-1em}
\end{figure*}

When people shoot a photograph with either a film-based camera or a digital camera, the scene in front of the camera is recorded on either an analog film or a digital sensor under the control of a shutter. 
An image is then formed by integrating all incoming photons on the film or sensor during the period of exposure. 
In this process, different shutter strategies bring different imaging effects.
When the shutter speed is slow and there are objects moving in the scene, a blurry image will be produced. When the shutter speed is fast but the shutter frequency is low, the fast-moving objects in the video will suffer from stuttering. When a row-by-row readout scheme is adopted in the exposure process, rolling shutter artifacts such as jello effect may appear. 

The shutter strategy applied to the shooting process can not be changed once a photo is taken.
When the captured photo is affected by shutter-related problems, post-capture image processing techniques are typically employed for high-quality image restoration.
There has been a mass of methods for blur removal~\cite{DeblurSurvey, DeblurGAN, DeepDeblur}, video frame interpolation~\cite{DeblurSurvey, SloMo, DAIN}, and rolling shutter correction~\cite{DSUN, JCD}. 
{
}
However, most existing works deal with these issues separately, with each model only addressing a specialized issue. 
In this work, we would like to explore a way for \textit{re-exposing the captured photo}, which is a more flexible method for addressing all aforementioned issues, and even their combinations.
Similar to the light field camera refocusing a photo after it has been captured, we aim at adjusting the camera's shutter (\eg, timestamp, speed, or type) in the post-processing to achieve re-exposure.

Inspired by the physical imaging process of a camera where an image is dependent on the visual content of the captured scene and the shutter applied,
we propose to re-expose the captured image to produce a new image with a desired shutter manner under a neural network based framework.
The first step is to project the visual content corresponding to the captured scene into a latent space, which works as the content source of re-exposure. 
Since the traditional camera captures scene content during the exposure period through integration, there is much confusion and information missing in the recorded photo. 
Recently popular bio-inspired dynamic vision sensors like event camera~\cite{EventCmaera-1, EventCamera-2} and spiking camera~\cite{SpikingCamera-1, SpikingCamera-2} are able to record almost continuous brightness change by firing signal asynchronously, providing an opportunity to correct confusion and make up for missing information. 
With help of dynamic vision sensors, it is possible to approximately reconstruct the whole scene
in the latent space. 
The `recording' of the scene is then determined by a neural shutter, which is similar to the camera shutter whose open period determines the content of the scene being recorded.
However, different from the physical imaging process, 
the re-exposure is achieved by aggregating information from visual latent content rather than integration of photons. 
In a film camera, a film records the visual content by performing slight chemical changes proportional to the amount of light and is then developed into a photograph.
While in the neural network based re-exposure framework, a neural film is proposed as the carrier to aggregate visual content, and is `developed' by a decoder into the final desired photo.

As shown in Fig.~\ref{fig:structure_overview}, the proposed re-exposure framework consists of three components, including an encoder for constructing the visual latent space, a re-exposure module, 
and a decoder for `developing' neural film into a desired image. To construct the visual latent space, original sensor data including frames and events are processed by a CNN encoder and a Bi-LSTM encoder respectively, producing pyramid representation as visual latent content. 
Then a re-exposure module is followed to aggregate information in visual latent content to neural film.
The re-exposure module is mainly composed of self-attention layers which promote interaction between neural film and visual latent content, and a cross-attention layer which aggregates information to neural film. Furthermore, temporal information is integrated into the computation of self-attention to promote interaction and aggregation, and a feature enhancement module is used to promote interaction within neighborhood. 
Once neural film is re-exposed, a decoder is applied to `develop' it into a desired image.

The proposed neural re-exposure framework is validated on several shutter-related tasks, including image deblur, video frame interpolation, and rolling shutter correction and combination of them. The proposed method, with a unified framework, is able to re-expose different kinds of input image(s) into a desired one by adjusting the neural shutter, as shown in Fig.~\ref{fig:teasing_figure}.

\section{Related Works}

%


\subsection{Motion Deblur}
Motion blur occurs when the object or camera moves at high speed during the exposure period. 
To deblur the images, some methods~\cite{kernel1,kernel2} model make estimation about blur kernel first and conduct deconvolution with the estimated kernel.
Some methods~\cite{DeepDeblur,MIMO,NAFNet} adopt the encoder-decoder architectures to deblur images with neural network. 
Due to the complexity of blur patterns and lack of motion information within the exposure period, the performance of these methods is still limited especially when it comes to scenes with complex motion.

Benefiting from the rich temporal information with the events, event-based methods  ~\cite{EDI,LEMD,LEDVDI,EVDI,E-CIR,REDNet} have achieved significant progress. 
Pan \etal~\cite{EDI} proposed the Event-based Double Integral (EDI) model by exploring the relationship between events, blurry images, and the latent sharp image to deblur the image by optimizing an energy function.
Considering the impact of noise and the unknown threshold of events, some methods ~\cite{LEMD,LEDVDI,EVDI} use deep learning networks to predict the sharp image based on the same principle. 
Song \etal~\cite{E-CIR} model the motion by means of per-pixel parametric polynomials with a deep learning model. 
REDNet \etal~\cite{REDNet} estimates the optical flow with the event to supervise the deblurring model with blurry consistency and photometric consistency. 
By investigating the impact of light on event noise, Zhou \etal~\cite{lowlight} attempted to estimate the blur kernel with events to deblur images by deconvolution. 
Sun \etal~\cite{EFNet} proposed a cross-modality channel-wise attention module to fuse event features and image features at multiple levels.


\subsection{Video Frame Interpolation}
Most frame-only methods~\cite{SloMo,AdaCoF,DAIN,RIFE} are based on linear motion assumption. These methods estimate the optical flow according to the difference between two frames, and linearly calculate the displacement from the key frames to the target timestamp.  Because of lack of motion information between frames.

Compared with frame-only interpolation, event-based interpolation methods are more effective due to the power of events in motion modeling. This makes them competent for scenarios with more complex motion patterns. 
Xu \etal~\cite{REDNet} proposed to predict optical flow between output frames to simulate nonlinear motion within exposure duration. 
He \etal~\cite{timereplay} proposed an unsupervised event-assisted video frame interpolation framework by cycling the predicted intermediate frames in extra rounds of frame interpolation.
Tulyakov \etal~\cite{TimeLens} designed a frame interpolation framework by combining a warping-based branch and a synthesis-based branch to fully exploit the advantage of fusion of frames and events. 

\subsection{Rolling Shutter Correction}
Rolling shutter effect is caused by the row-by-row readout scheme, in which each row of pixels is exposed at a different time. 
Frame-only unrolling is mostly based on the motion flow and linear motion assumption. 
Fan \etal~\cite{RSSR,CVR} proposed to estimate the motion field between two adjacent input rolling shutter images, and predict the global shutter image based on that. 
In SUNet~\cite{SUNet} and DSUN~\cite{DSUN} pyramidal cost volume is computed to predict motion field and global shutter image is predicted by warping features of key frames based on that.
Zhou \etal~\cite{EvUnroll} introduced the event data to the unrolling task, and designed a two-branch structure which fully leverages information with frames and events to correct the rolling shutter effect.

\subsection{Joint Tasks}
There have already been some efforts in dealing with multiple tasks simultaneously. 
Some methods~\cite{EVDI, LEDVDI, DeMFI} deal with image deblur and frame interpolation simultaneously.
DeMFI~\cite{DeMFI} takes blurry key frames as input, deblurring the image with a flow-guided module and interpolating sharp frames with a recursive boosting module. 
Zhang \etal~\cite{EVDI} and Lin \etal~\cite{LEDVDI} unifid the image deblur and frame interpolation with the help of events. 
EVDI~\cite{EVDI}  predicts sharp images of a given timestamp by leveraging blurry images and corresponding events, which are then fused as interpolation results. 
Lin \etal~\cite{LEDVDI} proposed to use events to estimate the residuals for the sharp frame restoration, and the restored frames compose a video of higher framerate.

Zhong \etal~\cite{JCD} and Zhou \etal~\cite{EvUnroll}  proposed methods to convert blurry rolling shutter images into sharp global shutter images. 
JCD~\cite{JCD} joint address motion blur and rolling shutter effect with a bi-directional warping stream and a middle deblurring stream.
EvUnroll~\cite{EvUnroll} is an event-based method that deblurs the blurry rolling shutter image first, then corrects the rolling shutter effects in a two-branch structure.

It is worth noting that, although above methods address multiple issues in a single model, they handle each aspect of the joint task with a corresponding module in a multi-stage manner.
In this work, we propose a unified framework to deal with all shutter-related problems. By re-exposing the captured image with a desired shutter, all aspects of the joint task can be addressed in a unified way.

\section{Problem Setting}
\label{sec:Problem Setting}

\begin{figure*}
	\centering
	\includegraphics[width=1.0\textwidth]{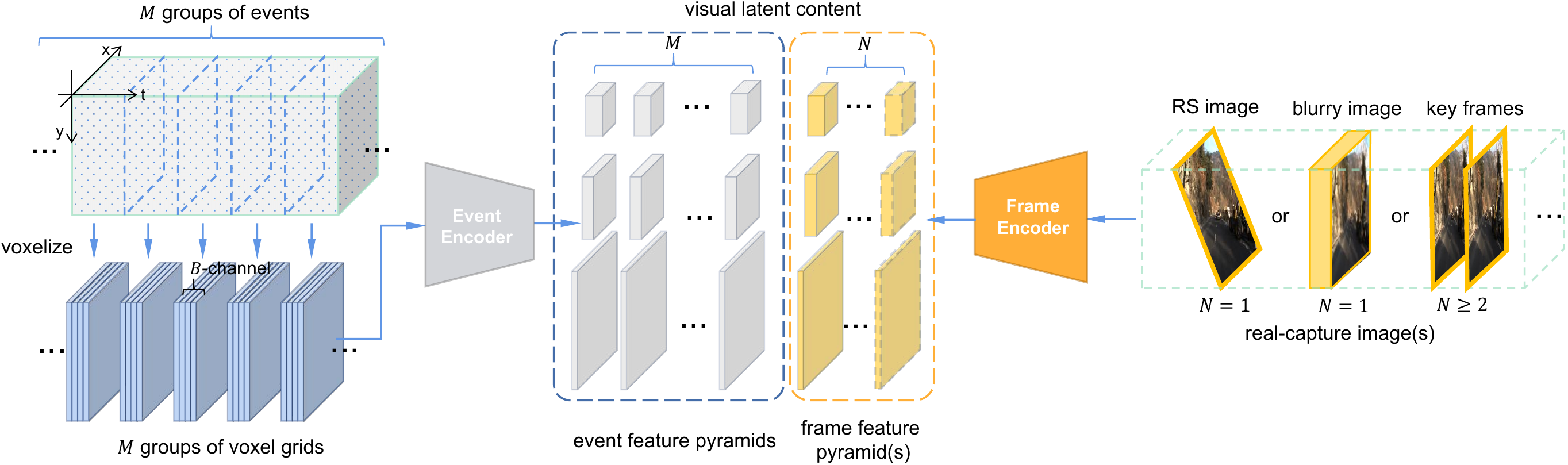}
	\caption{Illustration of the construction of visual latent content with events and real-captured frame(s).}
	\label{fig:visual_latent_space_construction}
	\vspace{-1.0em}
\end{figure*}


In the camera imaging process, the content of an image is determined by both the scene and the applied shutter:

\begin{equation}
I(x, y) = \int_{T}^{T+\Delta{T}} V(x,y,t)S(x,y,t) dt,
\label{equ:shutter_operation}
\end{equation}
where, $V(x,y,t) \in \mathbb{R}^3$ represents the visual content of the scene at position $(x, y)$ and time $t$, $[T, T+\Delta{T}]$ denotes a time interval, $I(x, y)$ denotes the captured image's pixel value at $(x,y)$, $S(\cdot)$ is the shutter function that controls the exposure process. 
\begin{equation}
S(x,y,t) = u(t-t_a(x,y)) - u(t-t_b(x,y)) 
\end{equation}
where $u(t)=\mathbbm{1}_{t>0}(t)$ is unit step function, and $t_a(x,y)$ and $t_b(x,y)$ respectively denote the start and end of exposure for pixel $(x,y)$, which may correspond to various shutter strategies, and they satisfy $T < t_a(x,y) < t_b(x,y) < T + \Delta{T}$.
For an image captured by global shutter, both $t_a(x,y)$ and $t_b(x,y)$ are constant for all positions, which can be represented as $t_a(x,y) = t_1$ and $t_b(x,y) = t_2$.
For an image captured by rolling shutter, 
$t_a(x,y)$ and $t_b(x,y)$ are constant for all columns but vary over rows, which can be represented as $t_a(x,y) = t_1 + \alpha y$ 
and $t_b(x,y) = t_2 + \alpha y$,
with 
$\alpha$ being readout delay between adjacent rows.
As for exposure duration $\Delta{t} = t_b(x,y) - t_a(x,y)$ which is constant across positions, the longer the exposure duration, the heavier the motion blur; while the shorter the duration, the sharper the captured image.
Content of a photo is fixed once it is taken; however, we would like to achieve image re-exposure in post-processing with help of event data.

By adjusting the shutter strategy to a desired type, speed, or timestamp, we can obtain a novel desired image.
Hence, we can have a unified framework to address multiple shutter-related image restoration tasks.
{\flushleft\textbf{Image deblur:}} given a blurry image and events triggered in the exposure period, images with different levels of sharpness could be obtained as the model's output by forcing 
$t'_b(x,y) - t'_a(x,y) = \delta$ (\eg ~$\delta = 0$), 
where $t'_a$ and $t'_b$ are for the re-exposed image, $\delta$ is the duration corresponding to the desired sharpness.
{\flushleft\textbf{Rolling shutter correction:}} given a rolling shutter (RS) image and the corresponding events, a global shutter (GS) image could be produced by forcing $t'_a(x,y) = t'_1$ and $t'_b(x,y) = t'_2$, where $t'_1$ and $t'_2$ correspond to the shutter time of the desired GS image.
{\flushleft\textbf{Video frame interpolation:}} given two frames at $t_1$ and $t_2$ and the corresponding events in-between, a series of intermediate frames could be predicted by forcing  $t'_{i,a}(x,y)=t'_{i,b}(x,y)=t'_i$, where $t'_i = \frac{i(t_2 - t_1)}{N}$, $i = 1, 2, \dots, N-1$.

Moreover, the proposed framework could deal with coupling of several issues and achieve image deblur simultaneously with RS correction or frame interpolation.

\section{Method}

In this section, we present Neural Image Re-Exposure (NIRE for short), 
a neural network based framework that could re-expose the captured scene to a new image with a desired shutter manner.
As shown in Fig.~\ref{fig:structure_overview}, the proposed framework consists of three modules, \ie, an encoder for constructing visual latent space, a re-exposure module, and a decoder for `developing' neural film into a desired image.

\subsection{Construction of Visual Latent Space}
\label{sec:Construction of Visual Latent Space}

\begin{figure*}[h!]
\centering
\includegraphics[width=1.0\textwidth]{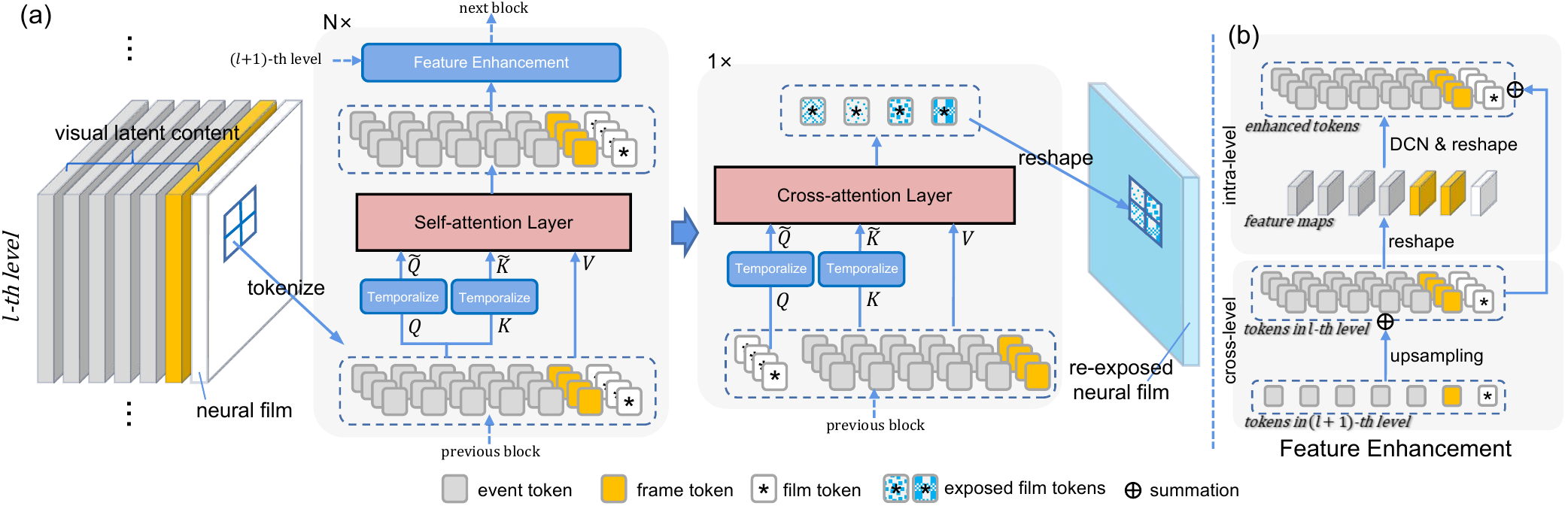}
\caption{Illustration of the re-exposure module
(a) shows the overall structure of the re-exposure module.
(b) shows the feature enhancement module.
}
\label{fig:exposing_simulation}
\vspace{-1.0em}
\end{figure*}

The original sensor data from synchronized event and conventional cameras are used to construct visual latent space
As for asynchronous and sparse event data $E = \{(x,y,p,t)\}, t \in [T,T+\Delta T]$, they are split into $M$ segments according to their time stamps to avoid too much compression over time, as shown in Fig.~\ref{fig:visual_latent_space_construction}. For each segment, voxel grid representation~\cite{voxel_grid} with $B$ bins is used to model spatial-temporal information with events, which is denoted as ${V_m}$.
We sequentially fed $\{V_m\}^{M}_{m=1}$ into an event encoder, obtaining $M$ feature pyramids with $m-$th being
$\mathcal{E}_m=\{ \mathcal{E}^l_m \}_{l=1}^{L}$, $\mathcal{E}^l_m \in \mathbb{R}^{C_l\times\frac{H}{2^{l-1}}\times\frac{W}{2^{l-1}}}$, $L$ is the total number of levels, and $C_l$ is the number of channels of the $l$-th level.
To fully gather information with events about brightness change in the interval, a bi-directional LSTM based encoder, which takes as input a sequence of voxel grids $\{V_m\}, m = 1, 2, ...,M$, is employed to aggregate information with events in different segments.
We follow the design scheme of the recurrent encoder in E2VID~\cite{E2VID} but add bi-directional propagation for promoting aggregation.

As for the captured $N$ images, each of them is processed by a fully convolutional multi-scale encoder 
(the first three stages of ResNet-34~\cite{ResNet} is used in this work) and a feature pyramid $\mathcal{I}_n=\{ \mathcal{I}^l_n \}_{l=1}^{L}, \mathcal{I}^l_n \in \mathbb{R}^{C_l\times\frac{H}{2^{l-1}}\times\frac{W}{2^{l-1}}}$ is obtained. 
Here the number of images $N$ depends on the considered task, \eg $N=2$ for the VFI task and $N=1$ for the image deblur task.
The combination of both event and frame feature pyramids composes the scene's representation on visual latent space, \ie visual latent content.

\subsection{Image Re-Exposure}
The re-exposure module takes as input visual latent content, neural film, and corresponding time encodings, and outputs a re-exposed film.
It is mainly composed of self-attention layers and a cross-attention layer as shown in Fig.~\ref{fig:exposing_simulation}.
By treating the representation of visual latent content and neural film as a set of tokens, these attention layers are optimized to aggregate information from visual latent content to the neural film for later `film development'.

\paragraph{Neural film.} 
The neural film works as an information aggregator in the image re-exposure process.
It has the same spatial shape as visual latent content's feature pyramid representation and is denoted as $\mathcal{X}_i=\{ \mathcal{X}^l_i \}^L_{l=1}$, with $\mathcal{X}^l_i \in \mathbb{R}^{C_l\times\frac{H}{2^{l-1}}\times\frac{W}{2^{l-1}}}$.
For each level in the pyramid $\mathcal{X}^l_i$, the `base color' of neural film is initialized as a set of learnable parameters, which are the same across all locations and do not change once training is over.
In a film camera, the emulsion would gradually experience chemical change during exposure to light in the scene.
Similarly, neural film, which is designed for collecting information from visual latent content, would gradually vary as it goes through successive neural network layers in the re-exposure module.

\paragraph{Interaction between neural film and visual latent content.} 
Several self-attention layers are employed to promote interaction between neural film and visual latent content.
For each level in the pyramid, event encoded feature maps, frame encoded feature maps and neural film are all divided into non-overlapping windows of the same size $r\times r$ similar to SwinTransformer~\cite{SwinTransformer}.
Each position in a window of a type of feature map is treated as a token and self-attention operations are computed over all tokens within each window of all types of feature maps.
In addition, temporal information is also integrated into the computation of self-attention to achieve time-aware re-exposure.

Each token is temporalized by adding an additional time encoding vector that represents the time range it spans $[t_a, t_b]$.
For example, as for event encoded feature maps corresponding to a certain segment of events, its time encoding should encode the start and end of that segment; as for a frame, its time encoding should encode the start and end of its exposure time; as for neural film, its time encoding at a certain position should encode the start and end of its expected exposure time. 
The encoding of a timestamp is represented in a similar way as the positional encoding of transformer~\cite{Transformer},
\begin{equation}
\resizebox{0.9\columnwidth}{!}{
    $\gamma(t)=\left(\sin \left(2^0 \pi t\right), \cos \left(2^0 \pi t\right), \cdots, \sin \left(2^{K-1} \pi t\right), \cos \left(2^{K-1} \pi t\right)\right)$
}
\label{equ:time_encoding}
\end{equation}
In practice, the time range of the whole considered events $[T,T+\Delta T]$ is normalized to $[0, 1]$, so we have $t \in [0, 1]$.
The encoding of time range is further represented as concatenation of encodings of both start and end timestamps of a certain range, 
\begin{equation}
	\mathcal{T}(t_a, t_b) = [\gamma(t_a(x, y)), \gamma(t_b(x, y))].
	\label{equ:time_encoding_pair}
\end{equation} 
Note that for RS-related frame tokens and film tokens, the start and end timestamps are dependent on their position.
Since the time encodings for film tokens could determine the visual content to be re-exposed, which is similar to the shutter in a camera, thus they are given the term neural shutter.

In the local self-attention layer, tokens within a window are projected to $d-$dimension queries $Q$, keys $K$ and values $V$ with three linear layers $f_Q$, $f_K$, and $f_V$, as shown in Eq.~\ref{equ:qkv}.
\begin{equation}
    [Q, K, V] = [f_Q(Z), f_K(Z), f_V(Z)]
    \label{equ:qkv}
\end{equation}
The time encodings are also projected to dimension $d$ by a linear layer $f_T$.
These tokens are temporalized by adding the projected time encodings to the original queries and keys, making them time-aware, as shown in Eq.~\ref{equ:temporalize}.
\begin{equation}
    \widetilde{Q}=Q + f_T(\mathcal{T}), \widetilde{K}=K + f_T(\mathcal{T}).
    \label{equ:temporalize}
\end{equation}
Then the output of self-attention layer could be computed as below,
\begin{equation}
    Attention(\widetilde{Q}, \widetilde{K}, V) = softmax(\widetilde{Q}\widetilde{K}^T/\sqrt{d})V.
\end{equation}
Layers such as Multi-Head attention, LayerNorm, and FFN layers, are kept the same as the original self-attention layer~\cite{Transformer}.
In this way, tokens of both visual latent content and neural film could interact sufficiently with each other by exchanging their own information.
In addition, a feature enhancement module is adopted to promote the interaction within neighborhood, where cross-level aggregation and a deformable layer\cite{DCN} are used to increase receptive field. Then they are reshaped back to a set of tokens. Hence interaction within a larger neighborhood is also promoted and more information could be leveraged in the aggregation.

\paragraph{Aggregation to neural film.} 
To aggregate visual information to the neural film which could be `developed' later, a cross-attention layer is added.
In the cross-attention layer, the operations are almost the same as the ones in self-attention layers, except that only tokens from the neural film are taken as queries and those from visual latent content are taken as keys and values, as shown below.
\begin{equation}
\resizebox{0.85\columnwidth}{!}{
$[Q, K, V] = [f_Q(Z^{\mathcal{X}}), f_K([Z^{\mathcal{E}}, Z^{\mathcal{I}}]), f_V([Z^{\mathcal{E}}, Z^{\mathcal{I}}])]$
}
\end{equation}

\subsection{Decoder for neural film `development'}
\label{sec:Decoder}
Similar to the last step of photography where an image is developed from the exposed film,
a decoder is used to `develop' the re-exposed neural film to a desired image.
The processed film tokens in the re-exposure step are first reshaped back to the shape of feature maps, so we can obtain a pyramid of the re-exposed neural film.
The pyramid is sent to the decoder one by one from the coarsest level to the finest level.
Each level is processed by a stack of convolutional layers and an upsampling layer. 
The upsampled feature is added to a finer level for further processing.
Finally, the decoder could give out a desired image corresponding to the original scene while controlled by the neural shutter.

\section{Experiments}

Since the proposed image re-exposure algorithm is able to deal with several shutter-related tasks within a unified framework, it is evaluated on tasks including image deblur, video frame interpolation (VFI), rolling shutter (RS) correction, and jointly deblurring and frame interpolation.

\subsection{Datasets}
Two datasets, GoPro~\cite{GoProDataset} and Gev-RS~\cite{EvUnroll}, are used for training and quantitative evaluation in our experiments. 
GoPro~\cite{GoProDataset} is a dataset consisting of sequences shot by a GoPro camera with a frame rate of 240 FPS and a resolution of 1,280$\times$720. 
It could provide training and test samples for tasks including image deblur~\cite{EFNet}~\cite{DeblurGAN}~\cite{SRN}, frame interpolation~\cite{TimeLens}~\cite{DAIN}~\cite{SloMo}, and jointly deblurring and frame interpolation~\cite{DeMFI}~\cite{TNTT}.
Gev-RS~\cite{EvUnroll} is a dataset collected for event-base rolling shutter correction. It is composed of sequences of 5,700 FPS frames recorded by high-speed Phantom VEO 640 camera such that high-quality RS images and event streams could be simulated. 
For each task, we follow its common evaluation setting for fair comparison.

\subsection{Training Strategy}
In this work, since the proposed image re-exposure method could unify several shutter-related tasks within one framework, a multi-task training strategy is adopted.
Specifically, for each batch, we randomly select a task and feed the corresponding frame(s) (\eg a blurry image, an RS image, or a pair of key frames), accompanying events and the neural shutter of a desired image into the model. The desired image is then used as groundtruth to supervise the training with a combination of Charbonnier loss~\cite{CharbonnierLoss} and perceptual loss~\cite{PerceptualLoss} 
as the target function. 
Since synthetic data is used for training and shutter speed for sharp image is very short, the exposure duration of the desired sharp GS image is ignored, \ie $t'_a(x,y)=t'_b(x,y)=t_{gt}$ where $t_{gt}$ can be an arbitrary timestamp of a sharp GS image used in synthesizing the blurry/RS image. 
In addition, since the proposed method can re-expose an image to any desired one with the corresponding neural shutter, we can also train the model to reconstruct the input frame.
In this case, the neural shutter is set identical to the time encodings of shutter time corresponding to the input image, \ie, $t'_a(x,y)=t_a(x,y), t'_b(x,y)=t_b(x,y)$.
The input images are randomly cropped into $128\times128$ patches in training. We train our model till 60,000 iterations with a batch size of 32 on a Tesla A100 GPU.

\begin{figure*}[t!]
	\centering
	\includegraphics[width=1.0\textwidth]{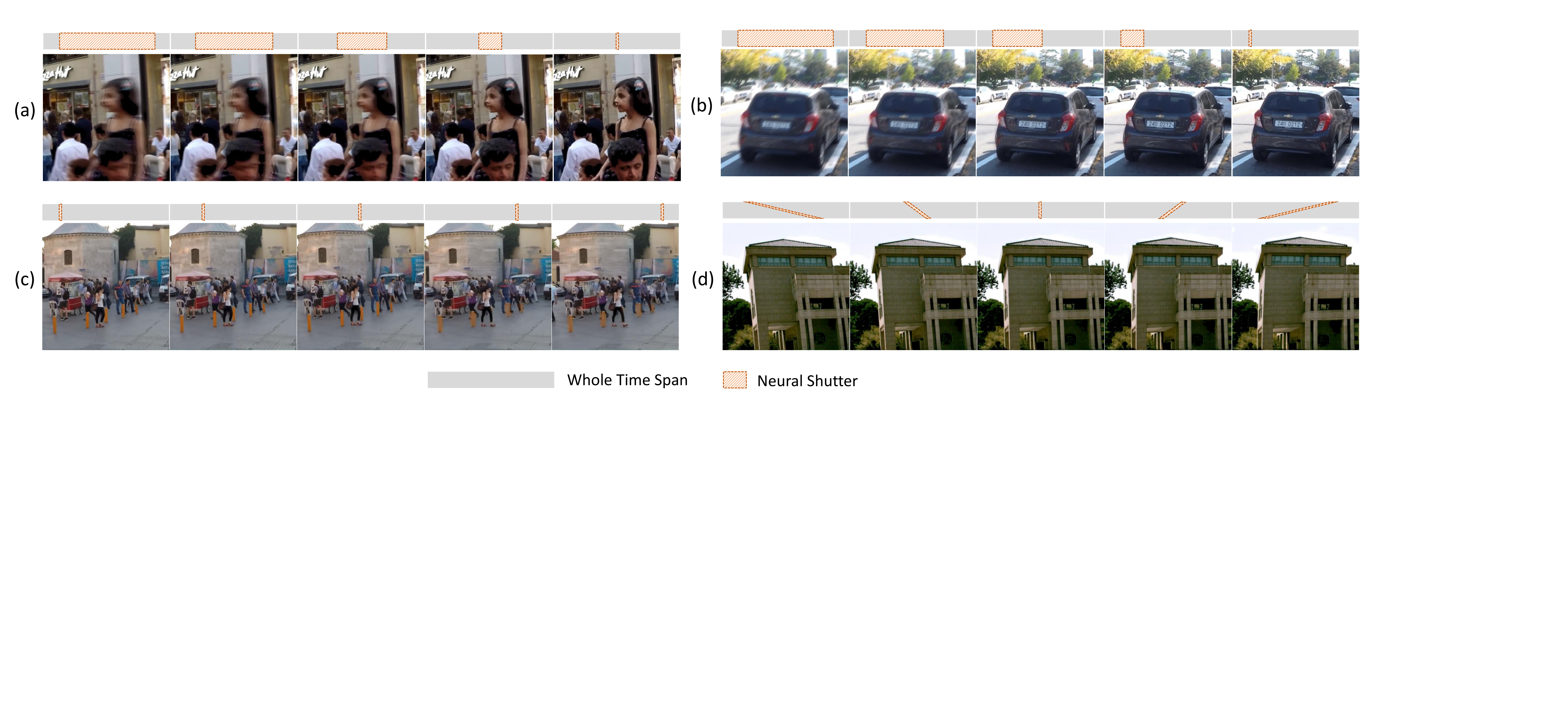}
	\caption{
		Qualitative results of the simulated shutter-controlled exposure.
		The quality and content of the frames are controlled by the neural shutter.
	}
	\label{fig:qualitative_results}
\end{figure*}

\subsection{Deblur}
Following the experiment setting in~\cite{EDI, EFNet}, the 3,214 blurry-sharp image pairs in GoPro dataset are split into 2,103 pairs for training and 1,111 pairs for testing. 
The blurred images are synthesized by averaging  consecutive high-framerate sharp frames.
Events are synthesized by applying a popular event camera simulator ESIM~\cite{ESIM-Simulator} to high-framerate sharp frames.
Hence under this setting, duty cycle is equal to 1, and start and end timestamps of exposure time are $t_a=0, t_b=1$. For quantitative comparison with SOTA methods, the desired shutter is $t'_a=t'_b=0.5$.
\begin{table}[b!]
\centering
\caption{Performance on image deblur.}
\resizebox{0.75\columnwidth}{!}{
\begin{tabular}{@{}lcccc@{}}
	\toprule
	\textbf{Methods}  			&\textbf{event}	& \textbf{PSNR} & \textbf{SSIM} \\
	\midrule
	E2VID~\cite{E2VID} 			&\ding{51}				& 15.22         & 0.651  	    \\
	DeblurGAN~\cite{DeblurGAN}	&\ding{55}				& 28.70         & 0.858         \\
	EDI~\cite{EDI}     			&\ding{51}				& 29.06         & 0.940         \\
	DeepDeblur~\cite{DeepDeblur}&\ding{55}				& 29.08         & 0.914         \\
	DeblurGAN-v2~\cite{Dbl-v2}	&\ding{55}				& 29.55         & 0.934         \\
	SRN~\cite{SRN}     			&\ding{55}				& 30.26         & 0.934         \\
	SRN+~\cite{SRN}   			&\ding{51}				& 31.02         & 0.936         \\
	DMPHN~\cite{DMPHN}          &\ding{55}			    & 31.20         & 0.940         \\
	D$^2$Nets~\cite{D2Net}		&\ding{51}				& 31.60         & 0.940         \\
	LEMD~\cite{LEMD}		    &\ding{51}				& 31.79         & 0.949         \\
	Suin et al.~\cite{SAPHNet}	&\ding{55}				& 31.85         & 0.948         \\
	SPAIR~\cite{SPAIR} 			&\ding{55}				& 32.06         & 0.953         \\
	MPRNet~\cite{MPRNet}		&\ding{55}				& 32.66         & 0.959         \\
	HINet~\cite{HINet} 			&\ding{55}				& 32.71         & 0.959         \\
	ERDNet~\cite{ERDNet}		&\ding{51}				& 32.99         & 0.935         \\
	HINet+~\cite{HINet}			&\ding{51}				& 33.69         & 0.961         \\
        NAFNet~\cite{NAFNet} &\ding{55} &33.69 &0.967 \\
	EFNet~\cite{EFNet}     			&\ding{51}				& \textbf{35.46} 		& 0.972         \\
	\midrule
	NIRE 						&\ding{51}				& 35.03         & \textbf{0.973}         \\

	\bottomrule
\end{tabular}
}
\label{tab:prec_deblur}
\end{table}

As shown in Tab.~\ref{tab:prec_deblur},
the proposed NIRE outperforms most frame-only and event-based deblur SOTA deblur methods, and achieves comparable performance with the latest event-based method EFNet.
This demonstrates the effectiveness and superiority of the proposed framework.
Most existing methods that once the model is trained it could only restore the sharp frame of a fixed timestamp (\eg middle of exposure time). 
However, the proposed method with a pretrained model is able to re-expose a blurry image to images of various sharpness and to images of different timestamps in the exposure period.
As shown in Fig.~\ref{fig:qualitative_results}, subfigure (a) produces several images of various sharpness with the same middle timestamp, and subfigure (b) produces similar ones but with the same start timestamp.   

\subsection{Video Frame Interpolation}
To validate the effectiveness of our method on VFI task,
we evaluate the proposed NIRE method following the same setting as event-based VFI method\cite{TimeLens} on GoPro. 
In this task, we assume both given frames are sharp GS images and we have $t_{1,a}=t_{1,b}=0$ and $t_{2,a}=t_{2,b}=1$.
As shown in Tab.~\ref{tab:prec_vfi_gopro}, NIRE achieves much better performance than conventional frame-only methods and is on par with the specially designed event-based VFI method TimeLens~\cite{TimeLens}.
It slightly surpasses TimeLens on 7-skip settings, while performing slightly worse in terms of PSNR on 15-skip settings.
Fig.~\ref{fig:qualitative_results}(c) gives an illustration about how intermediate frames are predicted by adjusting neural shutter.
\begin{table}[h!]
\caption{Performance on video frame interpolation}
\centering
\resizebox{1.0\columnwidth}{!}{
\begin{tabular}{lccllll}
	\toprule
	\multicolumn{1}{c}{\multirow{2}{*}{\textbf{Method}}} & \multirow{2}{*}{\textbf{frames}} & \multirow{2}{*}{\textbf{events}} & \multicolumn{2}{l}{\textbf{7 frames skip}}
	& \multicolumn{2}{l}{\textbf{15 frames skip}} \\ \cmidrule(l){4-7} 
	
	\multicolumn{1}{c}{} 	& 			& 			& PSNR  & SSIM  & PSNR  & SSIM   \\ 
	\midrule
	DAIN~\cite{DAIN}		&\ding{51} 	&\ding{55}	& 28.81 & 0.876 & 24.39 & 0.736  \\
	SuperSloMo~\cite{SloMo}	&\ding{51} 	&\ding{55}	& 28.98 & 0.875 & 24.38 & 0.747  \\
	RRIN~\cite{RRIN}		&\ding{51} 	&\ding{55}	& 28.96 & 0.876 & 24.32 & 0.749  \\
	BMBC~\cite{BMBC}		&\ding{51} 	&\ding{55}	& 29.08 & 0.875 & 23.68 & 0.736  \\
	E2VID~\cite{E2VID}		&\ding{55} 	&\ding{51} 	& 9.74  & 0.549 & 9.75  & 0.549  \\
	EDI~\cite{EDI}			&\ding{51} 	&\ding{51} 	& 18.79 & 0.670 & 17.45 & 0.603  \\
	TimeLens~\cite{TimeLens}&\ding{51} 	&\ding{51} 	& 34.81 & 0.959 & \textbf{33.21} & 0.942  \\
	\midrule
	NIRE 		&\ding{51} &\ding{51} 	& \textbf{34.97} & \textbf{0.964} & 32.85 & \textbf{0.945}  \\	
	\bottomrule
\end{tabular}
}
\label{tab:prec_vfi_gopro}
\end{table}

\subsection{Joint Deblur and Rolling Shutter Correction}

The proposed image re-exposure method is also validated on the RS correction task, following the experiment setting of~\cite{EvUnroll}.
In that work, a more challenging setting assumes that an RS frame is blurry with $t_a(x,y)=\frac{y}{H+\delta}$ and $t_b(x,y)=\frac{y+\delta}{H+\delta}$, where $H$ is height of the image and $\delta$ is exposure time of the blurry image, \ie, pixels in different rows of an RS image are exposed at different times but with the same duration.
As for the desired sharp GS image, we assume the shutter time is $t_a'(x,y)=t_b'(x,y)=\frac{0.5H}{H+\delta}$.
With temporalized tokens in the proposed method, the difference between rows is taken into account in the re-exposure module such that proper information about visual content could be leveraged and aggregated to the neural film. Hence a sharp GS image free of rolling shutter artifact could be produced.

\begin{table}[t!]
	\centering
	\caption{Performance on joint deblur and RS correction.}
	\resizebox{0.7\columnwidth}{!}{
		\begin{tabular}{lccc}
			\toprule
			\textbf{Methods}  	& \textbf{events} 	& \textbf{PSNR}  	& \textbf{SSIM} \\	
			\midrule
			DSUN~\cite{DSUN}	& \ding{55}	  	& 23.10 			& 0.70 \\ 				
			JCD~\cite{JCD}      & \ding{55}	  		& 24.90 			& 0.82 \\				
			EvUnroll~\cite{EvUnroll}& \ding{51}     & \textbf{30.14} 			& \textbf{0.91} \\				
			\midrule
			NIRE     			& \ding{51}      	& 29.86 			& \textbf{0.91} \\				
			\bottomrule
		\end{tabular}
	}
	\label{tab:prec_unroll_gevrs}
\end{table}

As shown in Tab.~\ref{tab:prec_unroll_gevrs},
the method outperforms the frame-only methods and achieves comparable performance with the SOTA event-based method EvUnroll~\cite{EvUnroll},
demonstrating the effectiveness of the proposed NIRE framework in simultaneously removing rolling shutter artifact and blur.
Note that the method EvUnroll is a two-stage method with a deblur module specially for deblurring. And it is specially designed for the
event-based rolling shutter correction task, whose final result is a combination of the output of its synthesis branch
and warping branch.

\subsection{Joint Deblur and Frame Interpolation}
In addition, the proposed method is also validated on the task of joint deblur and frame interpolation following the same setting as~\cite{DeMFI}. The VFI task usually assumes the given key frames are sharp, however, frames are easily degraded by camera or object motion blur, which increases the difficulty of the VFI task.

\begin{table}[t!]
\centering
\caption{Performance on joint deblur and frame interpolation.}
\resizebox{0.9\columnwidth}{!}{
\begin{tabular}{@{}lllll@{}}
	\toprule
	\textbf{Methods}          	& \textbf{unified}  & \textbf{events}   & \textbf{PSNR}  & \textbf{SSIM}   \\
	\midrule
	%
	SRN~\cite{SRN} + SloMo~\cite{SloMo}      &\ding{55}		&\ding{55} 			& 24.72 & 0.7604 \\  
	SRN + MEMC-Net~\cite{MEMC-Net}     		&\ding{55}		&\ding{55} 			& 25.70 & 0.7792 \\  
	SRN + DAIN~\cite{DAIN}       			&\ding{55}		&\ding{55} 			& 25.17 & 0.7708 \\  
	EDVR~\cite{EDVR} + SloMo	&\ding{55}		&\ding{55} 			& 24.85 & 0.7762 \\  
	EDVR + MEMC-Net    			&\ding{55}		&\ding{55} 			& 27.12 & 0.8301 \\  
	EDVR + DAIN      			&\ding{55}		&\ding{55} 			& 29.01 & 0.8981 \\  
UTI-VFI          			&\ding{51}		&\ding{55} 			& 25.63 & 0.8148 \\	 
 	PRF~\cite{PRF}     			&\ding{51}		&\ding{55} 			& 25.68 & 0.8053 \\  
	TNTT~\cite{TNTT}   			&\ding{51}		&\ding{55} 			& 26.68 & 0.8148 \\  
	DeMFI-Net~\cite{DeMFI}		&\ding{51}		&\ding{55} 			& 31.25 & 0.9102 \\  
	\midrule
	NIRE-cascade     			&\ding{55}		&\ding{51} 			& 30.18 & 0.8923 \\
	NIRE		       			&\ding{51}		&\ding{51} 			& \textbf{33.43} & \textbf{0.9477} \\
	\bottomrule
\end{tabular}
\label{tab:prec_jdfi_gopro}
}
\end{table}

Simply cascading an image deblur model and a VFI model would lead to error accumulation and suboptimal performance.
While the proposed NIRE method could naturally handle such joint task by simply considering the shutter time of given key frames as $t_{1,b} - t_{1,a} = t_{2,b} - t_{2,a} = \delta$ with $\delta \neq 0$.
As shown in Tab.~\ref{tab:prec_jdfi_gopro}, NIRE outperforms existing frame-only methods by a large margin, 
showing its advantage in handling the joint task.
We also tried a cascaded variant of the proposed method by applying NIRE twice, denoted as NIRE-cascade.
It achieves significantly worse performance than addressing them in a unified manner, showing the advantage of end-to-end processing.

\subsection{Real-World Results}
\begin{figure}[h!]
	\centering
	\includegraphics[width=1.0\columnwidth]{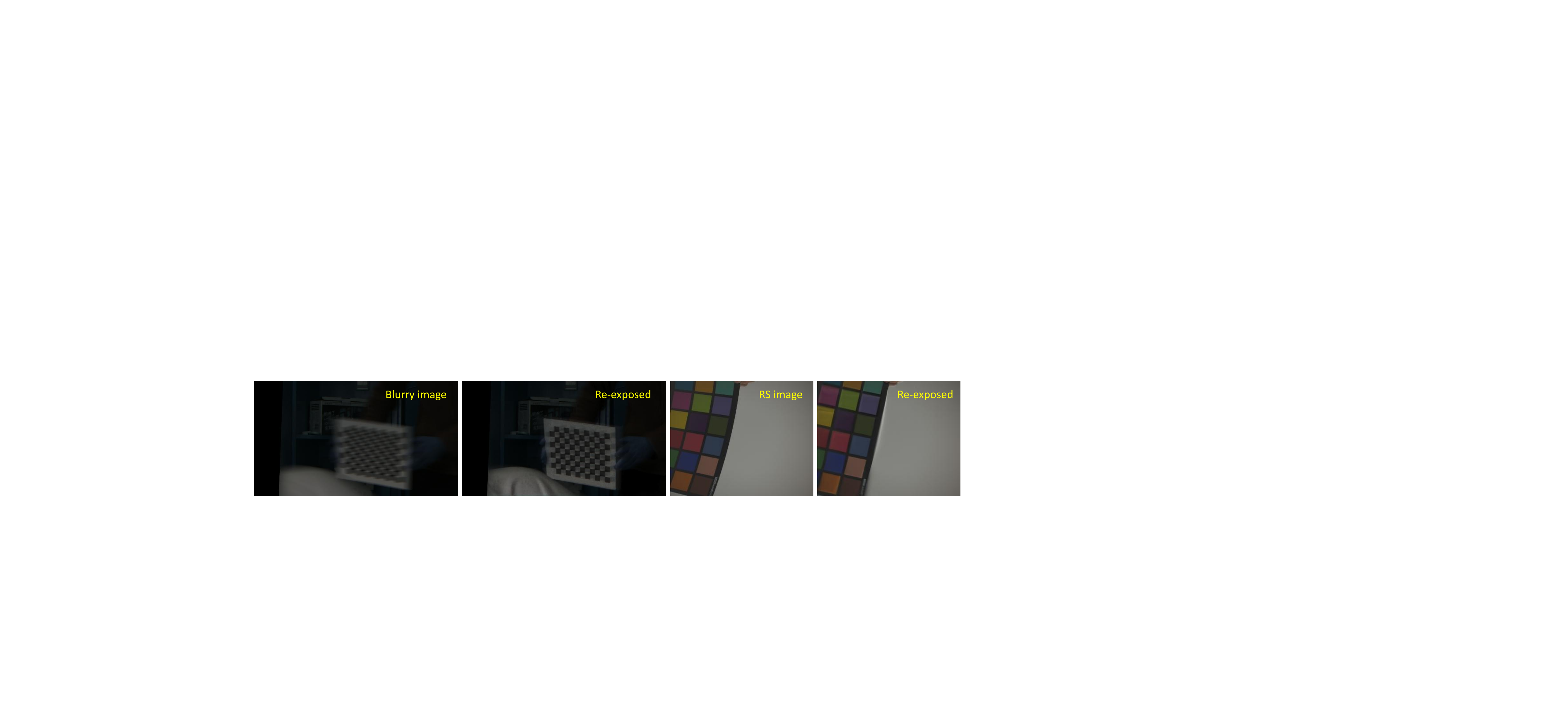}
	\caption{Qualitative result on real-captured data. }
	\label{fig:qualitative-real}
\end{figure}

We also evaluate the proposed method on real-captured frames and events. 
Zhou \etal~\cite{EvUnroll} collect some real-captured RS-event pairs with a designed hybrid camera system composed of a LUCID TRI054S IMX490 machine vision camera and a Prophesee Gen4.0 event camera which share the same field of view through a beam splitter.
In addition, we also collect some real blurry-event pairs for qualitative real-world deblurring evaluation. The capturing system is composed of a  Prophesee Gen4.0 event camera, a FLIR Blackfly and a beam splitter.
As shown in Fig.~\ref{fig:qualitative-real}, the proposed method, which is trained on synthetic data without any finetuning, performs reasonably well on real-captured data, showing its good generalization.

\subsection{Ablation Study}
Ablation study is conducted to investigate importance of components of the proposed framework.
In Tab.~\ref{tab:ablation}, NIRE w/o event represents the baseline with frames only, NTIRE w/o TimEnc denotes one by simply disabling the time encodings for visual latent content, NIRE w/o FeatEnhance denotes the one without feature enhancement module.
The results show the importance of event modality in compensating missing information, the role of time encodings in interaction among different kinds of information, and that of feature enhancement module in information aggregation.

\begin{table}[h!]
\caption{Ablation study of NIRE (in PSNR/SSIM).}
\resizebox{1.0\columnwidth}{!}{
\begin{tabular}{ccccc}
\toprule
Tasks     & VFI         & Deblur      & Unroll      & Deblur+VFI  \\ 
\hline
NIRE    & 34.97/0.964 & 35.03/0.973 & 29.86/0.908 & 33.43/0.948 \\
w/o Event & 30.40/0.886 & 29.53/0.928 & 24.08/0.803 & 26.46/0.815 \\
w/o TimEnc & 31.23/0.921 & 33.44/0.955 & 20.38/0.584 & 29.76/0.874 \\
w/o FeatEnhance   & 32.83/0.928 & 33.78/0.952 & 26.42/0.835 & 30.62/0.894 \\ 
\bottomrule
\end{tabular}}
\label{tab:ablation}
\end{table}

In addition, we compare the proposed multi-task training strategy with independent model training for different tasks. Tab.~\ref{tab:spec-vs-vers} shows that our method is not only parameter efficient in having a unified model dealing with several tasks simultaneously, but also performs on-par with or even better than independently trained counterparts.

\begin{table}[h!]
	\centering
	\caption{Comparison of specialized and versatile NIRE (in PSNR/SSIM).}
	\resizebox{1.0\columnwidth}{!}{
	\begin{tabular}{c|cccc}
		\toprule
		\diagbox{task}{strategy}& MT          & VFI         & Deblur      & Unroll      \\
		\hline
		VFI           & 34.97/0.964 & 34.44/0.955 & -           & -           \\
		Deblur        & 35.03/0.973 & -           & 34.72/0.966 & -           \\
		Unroll        & 30.08/0.909 & -           & -           & 30.04/0.909 \\
		\bottomrule  
	\end{tabular}
	\label{tab:spec-vs-vers}
	}
\end{table}


\section{Conclusion}

In this work, we propose a unified neural network based image re-exposure framework, which is able to re-expose a captured photo in the post-processing by adjusting the shutter strategy.
Event data is leveraged to compensate for information confusion and missing with frames.
A re-exposure module composed of attention layers is proposed to aggregate visual information to neural film under the control of neural shutter. 
The proposed unified framework is able to deal with multiple shutter-related tasks simultaneously with state-of-the-art performance.



{\small
\bibliographystyle{ieee_fullname}
\bibliography{egbib}
}

\end{document}